\documentclass[letterpaper, 10 pt, conference]{ieeeconf}  
\IEEEoverridecommandlockouts
\usepackage{graphicx}
\usepackage{mathptmx} 
\usepackage{times} 
\usepackage{amsmath} 
\usepackage{amssymb}  
\usepackage{units}
\usepackage[caption=false,font=scriptsize,justification=raggedright,singlelinecheck=true]{subfig} 
\usepackage{verbatim}  
\usepackage{nicefrac}		
\usepackage{acronym}		
\usepackage{glossaries}		
\usepackage[usenames,dvipsnames]{xcolor}			
\usepackage{cite} 
\usepackage{url} 
\usepackage[multidot]{grffile} 
\usepackage{multirow}
\usepackage{booktabs}
\makeatletter
\let\NAT@parse\undefined
\makeatother
\usepackage[ colorlinks=true, citecolor=black, linkcolor=black, urlcolor=black,  bookmarks=true, bookmarksnumbered=true, pdftex, plainpages=false, pdfpagelabels]{hyperref} 

\newacronym{ACFR}{ACFR}{Australian Centre for Field Robotics}
\newacronym{USyd}{USyd}{the University of Sydney}
\newacronym{AUV}{AUV}{autonomous underwater vehicle}
\newacronym{UAV}{UAV}{unmanned aerial vehicle}
\newacronym{SLAM}{SLAM}{simultaneous localisation and mapping}
\newacronym{SfM}{SfM}{structure-from-motion}
\newacronym{SNR}{SNR}{signal-to-noise ratio}
\newacronym{DFT}{DFT}{discrete Fourier transform}
\newacronym{FFT}{FFT}{fast Fourier transform}
\newacronym{SIFT}{SIFT}{scale invariant feature transform}
\newacronym{TP}{TP}{true positive}
\newacronym{FP}{FP}{false positive}

\title{\LARGE \bf
Burst Imaging for Light-Constrained Structure-From-Motion
}

\author{Ahalya Ravendran, Mitch Bryson, Donald G. Dansereau
\thanks{The authors are with the Australian Centre for Field Robotics (ACFR), School of Aerospace, Mechanical and Mechatronic Engineering, The University of Sydney and with the Sydney Institute for Robotics and Intelligent Systems, 2006 NSW, Australia.
{\tt\small arav3215, mitch.bryson, donald.dansereau@sydney.edu.au}}%
}

\newcommand{\myquat}[1]{\bar{#1}}

\newcommand{\q}{\myquat{q}}

\newcommand{\Cq}[2]{\boldsymbol{C}(\q)}

\newcommand{\Table}[1]{Tab. #1}

\begin{document}
\maketitle
\thispagestyle{empty}
\pagestyle{empty}

\begin{abstract}
Images captured under extremely low light conditions are noise-limited, which can cause existing robotic vision algorithms to fail. In this paper we develop an image processing technique for aiding 3D reconstruction from images acquired in low light conditions. Our technique, based on burst photography, uses direct methods for image registration within bursts of short exposure time images to improve the robustness and accuracy of feature-based \gls{SfM}. We demonstrate improved SfM performance in challenging light-constrained scenes, including quantitative evaluations that show improved feature performance and camera pose estimates. Additionally, we show that our method converges more frequently to correct reconstructions than the state-of-the-art. Our method is a significant step towards allowing robots to operate in low light conditions, with potential applications to robots operating in environments such as underground mines and night time operation.
\end{abstract}

\section{INTRODUCTION}
\label{sec:intro}

Current and emerging robots use vision sensors for a broad range of tasks including \gls{SLAM}, navigation, pose estimation, depth estimation and 3D reconstruction. State-of-the-art methods for structure-from-motion (SfM) and vision-based reconstruction (e.g.~\cite{Schonberger2016, Han2021}) perform well under good lighting conditions but fail to reconstruct in low light for tasks such as autonomous driving, drone surveillance and underground mining. When using light-constrained images, state-of-the-art methods~\cite{Schonberger2016} yield incorrect 3D shape estimation, inaccurate camera trajectories and in some scenes even fail to converge for light-constrained \gls{SfM}. This results from low \gls{SNR} in images, where true features are not detected and/or spurious features are detected and matched with features in other images for reconstruction. 
    
Burst photography is an established mobile photography technique which uses a series of consecutive frames with small exposure time to produce a single image with an improved \gls{SNR} upon merging~\cite{Liu2014, Hasinoff2016, Liba2019, Wronski2019}. Burst imaging has been shown to improve the \gls{SNR} in images without the need for additional ambient light sources~\cite{Hasinoff2016}. However, prior works and on-going developments on burst imaging are steered towards mobile photography, where the primary objective is to produce convincing content for human visual perception while accounting for camera motion due to handshake and scene motion within a single burst.
    
\begin{figure}[t]
	\centering
	\includegraphics[width=\columnwidth]{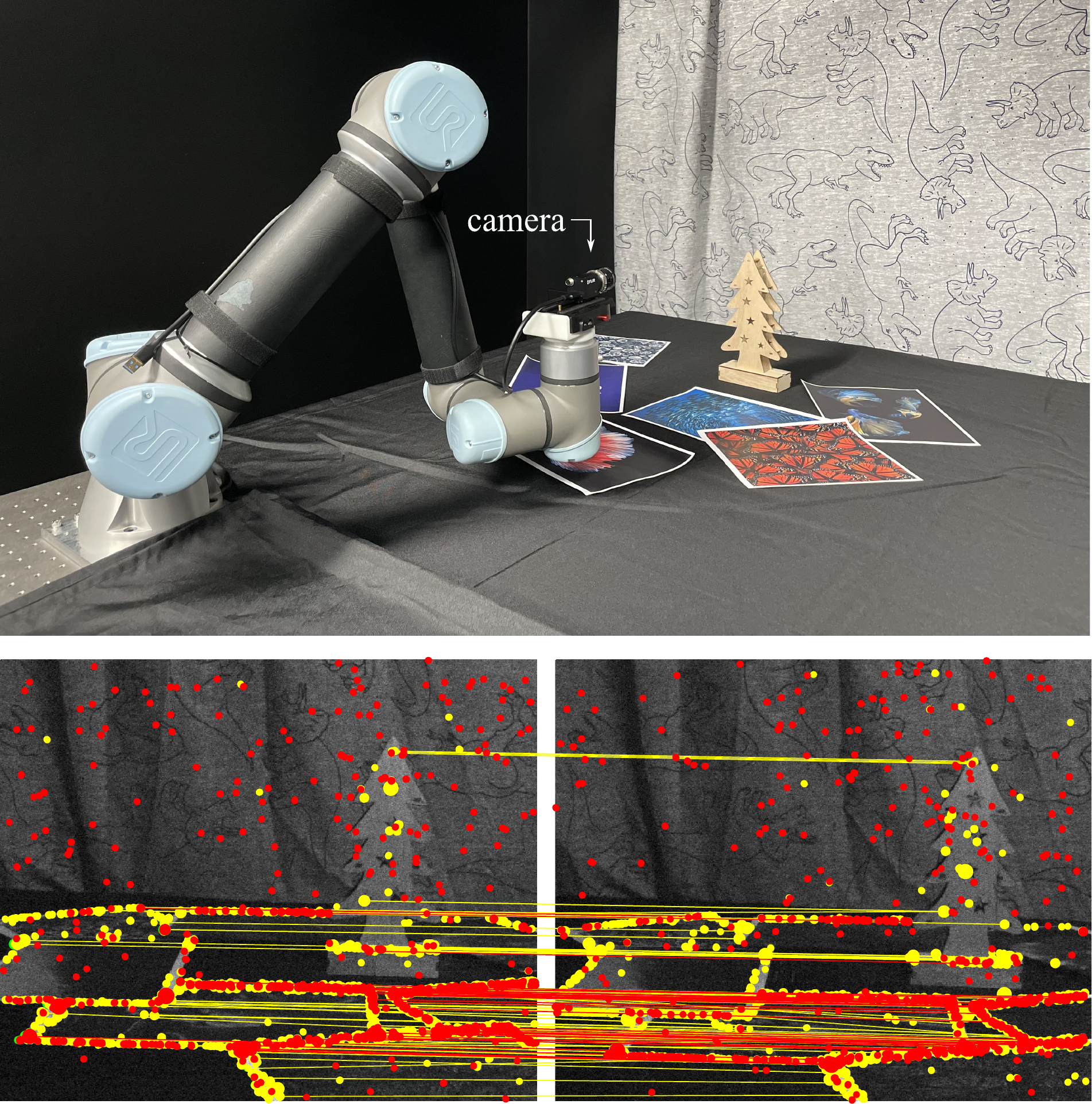}
	\caption{Visualisation of light-constrained scene. (top) Robot arm-mounted machine vision camera; (bottom) yellow: 110 matching pairs using our proposed merging approach; red: 43 matching pairs between consecutive conventional noisy images which fails to converge for reconstruction. It is important to note that 36\% of input images of this particular scene fails to find any matches in conventional noisy approach where as all input images are registered using our method for reconstruction. Our approach provides more stable set of true features and less spurious features for motion dependent applications such as SfM}
	\label{fig_1}
\end{figure}

We propose to reconstruct light-constrained scenes by adapting burst imaging for reconstruction. We describe an imaging pipeline that captures multiple bursts of images and accounts for motion variation within each burst using hierarchical tile-based alignment. We temporally merge each aligned burst to a single image using a voting scheme and then use these merged images in feature-based SfM to reconstruct scenes (see Fig.~\ref{fig_1}). We also consider adding spatial filtering separately on temporally denoised images to evaluate hybrid denoising.
    
Our key contributions are:
\begin{itemize}
  \item We establish the viability of using burst imaging to improve robotic vision in low light, and provide a set of recommendations for adopting this approach in reconstruction tasks such as \gls{SfM};
  \item We evaluate different approaches to burst imaging in robotics applications and show that burst capture with merge offers significant advantages in both computational requirements and performance; we offer a theoretical explanation and experimental evidence showing why this is so; and
  \item We demonstrate the proposed method improving low-light \gls{SfM} by yielding more precise 3D points, more true features, fewer spurious features, more precise camera trajectories and an ability to operate in lower light than was previously possible.
\end{itemize}

Our approach assumes that the level of noise present in images is not so great as to make information unrecoverable (e.g. the underlying signal being suppressed by quantization noise). Additionally, we assume that there exists sufficient overlap between subsequent sets of images bursts for registration, which is commonly available for mobile platforms moving at moderate speeds relative to their environment.
    
To validate our method, we mounted a monocular machine vision camera on a UR5e robotic arm, and captured burst imagery of real scenes in an illumination-controlled environment over different exposure times. The images are captured over 20 scenes with objects of various size, shape and colour as 22 bursts for each scene, yielding 6160 images in 880 bursts. We are releasing a dataset containing raw Bayer images, captured over the exposure time of 1ms and 0.1ms for a repeated trajectory. Code and dataset are available at \url{https://roboticimaging.org/Projects/BurstSfM/}. 
    
To evaluate, we compare our results against existing alternative approaches depicted in Fig.~\ref{fig_2}: burst without merge which includes every image in each burst, and conventional single-image capture. We demonstrate our proposed method outperforms the alternative approaches by reconstructing more true features with high feature localization accuracy and less spurious features as shown in Fig.~\ref{fig_1}. We also show our method yields accurate estimation of camera pose trajectory while minimising the failure cases of convergence compared to the existing alternative approaches.
    
This work opens the way for a broad range of applications in which low light commonly complicates vision such as autonomous driving and delivery drones. 
\section{RELATED WORK}
\label{sec:related}

When optimising camera exposure settings for low \gls{SNR} images, there exist many trade-offs in the final image quality. Conventional cameras can increase \gls{SNR} by either widening the aperture which reduces depth of field or increasing the exposure time, which for dynamic scenes, increases motion blur. 

\begin{figure}
	\centering
	\includegraphics[width=\columnwidth]{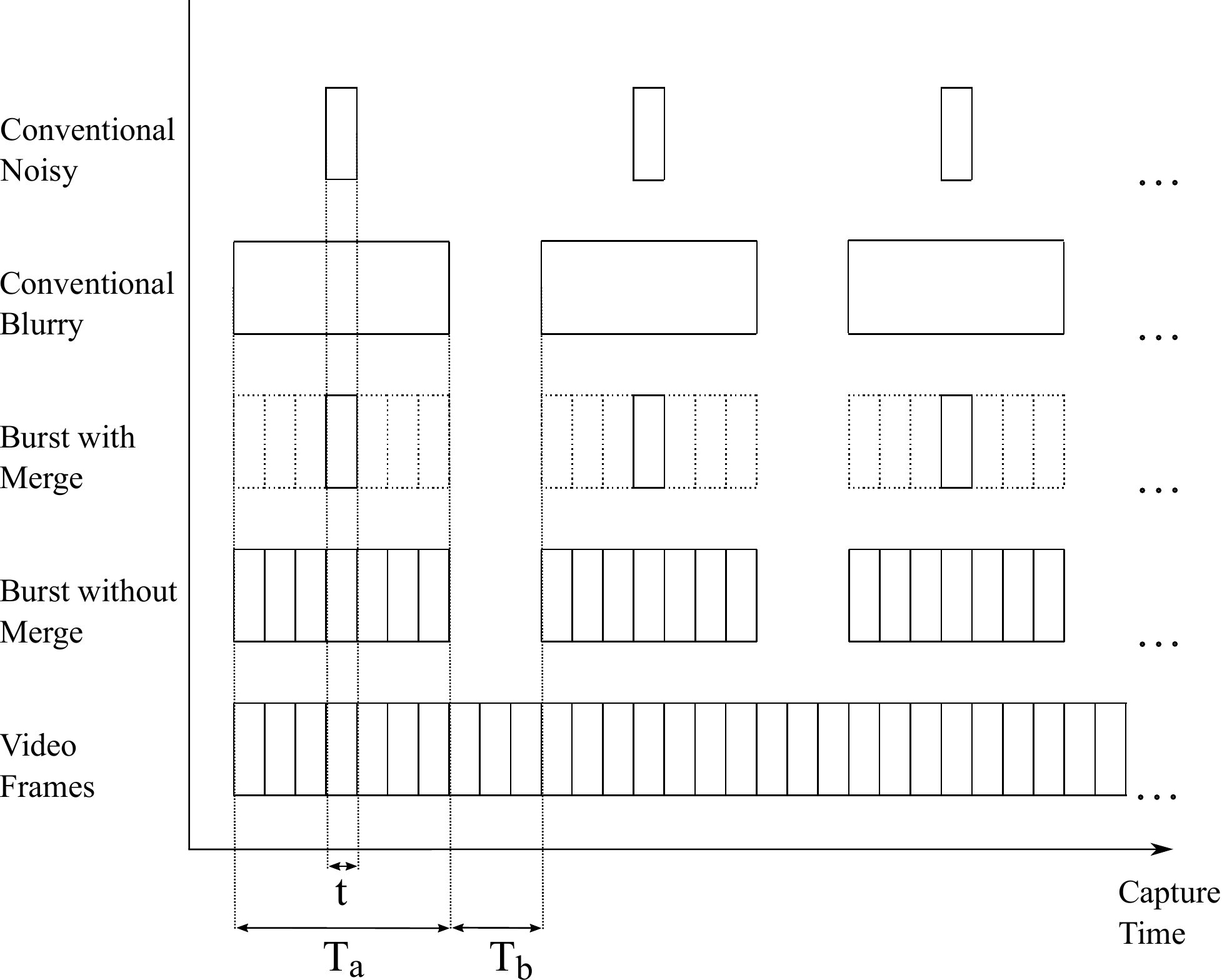}
	\caption{Alternative exposure schemes: Capturing images as a sequence/in video includes most of the information in the scene, but takes longer time to process. Burst without merge capture fewer images than video giving advantage over the time at the expense of information; Proposed approach, merge images in a burst ideally to the middle frame for robotic applications, to address for the relative motion between frames, improving the information over an intermediate time. Alternatively, conventional blurry images capture long exposure images with less noise but the quality of images are affected by motion blur; Conventional noisy images capture images trading-off motion blur for noise. Note, $T_a$ is the time taken for capturing $N$ number of the images in a burst. $t$ is the time taken to capture a single frame and $T_b$ is time taken between bursts}
	\label{fig_2}
\end{figure}

Considering alternative exposure schemes, depicted in Fig.~\ref{fig_2}, capturing images as a sequence over a long period of time yields a video. However, it is computationally intensive to process images at full video rates and in low light, each of these frames have low \gls{SNR}.

A number of existing computational imaging approaches successfully break discussed imaging trade-offs and take advantages of the alternative exposure schemes as demonstrated in coded aperture~\cite{Levin2007}, flexible depth of field photography~\cite{Nagahara2008}, flutter shutter~\cite{Raskar2006} and motion-invariant photography~\cite{Levin2008}. Unfortunately they require hardware modification to imaging sensors/cameras whereas our focus is on using existing monocular cameras to get quality images in low light for robotic vision. 

Learning-based techniques show promising results for denoising~\cite{Quan2020,Tassano2020,Yue2020}, deblurring~\cite{Nan2020,Pan2020,Yuan2020}, dehazing~\cite{Dong2020,Li2019,Liu2019} and image enhancement~\cite{Chen2018,Sharma2018,Wang2019}. However, they are not shown to generalize outside the training domain~\cite{Wei2020} and need large datasets and long training times. There is a factor of uncertainty~\cite{BarredoArrieta2020, Carvalho2019} in terms of interpretability and transparency, which imposes additional constraints on the use of deep learning approaches in robotics applications. Many of these techniques are also trained on simulated images rather than captured images and thus, often yield unreliable results for previously unseen terrain and environmental conditions~\cite{Ibarz2021,Choi2021}. 

Single-image based approaches are fundamentally limited by the amount of information in a conventional single image. They also have a tendency to produce visually pleasing results and not necessarily accurate information through learned image priors~\cite{Bhat2021}. This affects their reliability for applications such as reconstruction and tracking.

Hasinoff et al.~\cite{Hasinoff2010} exploit the time slice advantage and use burst imaging as a solution for night photography by capturing multiple underexposed images and merging them to get a single higher \gls{SNR} image. This is implemented as night sight in Google Pixel mobile phones~\cite{Hasinoff2016}, astrophotography mode~\cite{Liba2019}, multi-frame super resolution~\cite{Wronski2019} and using learning approaches~\cite{Mildenhall2018}. It has also been adapted as quanta burst photography~\cite{Ma2020} for single-photon sensors to produce high quality images. This and other previous work in mobile photography has focused mainly on producing visually pleasing still images from bursts with little camera motion. In this work we adapt burst imaging for 3D reconstruction with extended camera motion in low-light scenes.

Our work also considers considerably more challenging imagery than what is typically employed in existing burst photography work. Noisy images that are captured as a burst in mobile photography applications still have acceptable \gls{SNR} regimes for human visual systems and thus, the robots have the ability to perform reconstruction using existing state-of-the-art methods~\cite{Schonberger2016}. However, robots need the ability to visualize beyond this \gls{SNR} regime in operating environments such as mining quarry. As reconstruction in robotics has a wide coverage with large motions within a single trajectory unlike prior work which focuses on a burst, an efficient way of using multiple bursts for 3D image-based reconstruction applications is also an open question.

In this work, we adapt burst imaging for 3D reconstruction in low light. We do this by combining feature-based \gls{SfM} and burst photography, exploiting the advantages of each. By capturing rapid successions of frames, we enable the use of direct methods for image registration~\cite{Engel2014}, exploiting the small camera motion between frames to yield a strong \gls{SNR} advantage with moderate computational expense. After extracting features from the merged burst images, we apply feature-based \gls{SfM}~\cite{Schonberger2016} to handle large camera motions between bursts. Feature-based methods handle large camera motions~\cite{Mur2015} while benefiting from the improved feature quality associated with the burst-merged images.

Because our method compresses each burst into a single denoised image, total computation is lower during reconstruction compared with an approach employing all measured frames. Furthermore, the higher \gls{SNR} in our imagery can reduce the numbers of spurious features, again lowering computational requirements during reconstruction as there are fewer outliers to detect and reject.
\section{BURST-BASED SFM}
\label{sec:methods}

We capture multiple images as a burst and perform hierarchical tile-based alignment to compute the dense correspondence between every image in a single burst stack to a common image. We temporally merge the aligned stack with a voting scheme~\cite{Hasinoff2016} to get one denoised image per burst, and we evaluate different means of obtaining it: temporal, temporal with Wiener filtering~\cite{Hasinoff2016} and temporal with bilateral filtering. In the following sub-sections we outline how we capture bursts of images, align and merge each to a common image and use reconstruction pipeline.

\subsection{Image Acquisition for Reconstruction}
We capture $N$ frames over time $T_a$, before ceasing image acquisition for time $T_b$, and repeating this process, where $t$ is the time taken for capturing single image as in Fig.~\ref{fig_2}. We consider $N$ frames captured over time $T_a$ as a single burst and capture multiple bursts for a single trajectory. 
    
Capturing a greater number of images per burst (i.e. larger $T_a$) increases the resulting \gls{SNR} in merged images, provided there is a sufficient overlap. However, more images are also more computationally expensive for capturing, buffering and processing, so there is a trade-off in selecting the number of images in a burst between quality and computation. This is application-dependent: in our experiments, we use bursts of N=7 frames based on empirical evaluation.
    
\subsection{ Hierarchical Tile Alignment}
To address non-uniform apparent motion between frames in a burst, we perform coarse-to-fine alignment on multi level Gaussian pyramids of single channel images~\cite{Hasinoff2016}. Using the initial estimates from the coarser scale, we compute pairwise-tile-based alignment at each pyramid level by minimizing the distance measure between the common tile and the corresponding candidate tile of each alternate frame within a burst, similar to the approach in~\cite{Hasinoff2016}.

Directly minimizing an image-based misregistration measure based on pixel intensity errors is ideal here, because we capture consecutive images with a constant exposure and illumination, which is unlikely to change quickly over the duration of a single burst. To address illumination changes between bursts in a single trajectory, we use feature-based reconstruction.

We balance computational effort and the quality of alignment by fine tuning our alignment for different requirements that are common in reconstruction such as the number of channels in an input image, number of levels in an image pyramid and size of the input images in each pyramid level. Having high values for the aforementioned factors increases the computation at alignment. The number of levels in an image pyramid is also limited by the size of an input image.
    
There is a trade-off between the number of pyramid levels and the motion between the frames. If there is a larger motion between frames, we need larger search distances which require more pyramid levels or larger down sampling factors of input images on pyramid levels. This will reduce the likelihood of the search getting trapped in a local minimum and allow large motions between images. We empirically select the aforementioned factors to balance misalignment and motion between the frames.

\begin{figure}
	\centering
	\includegraphics[width=\columnwidth]{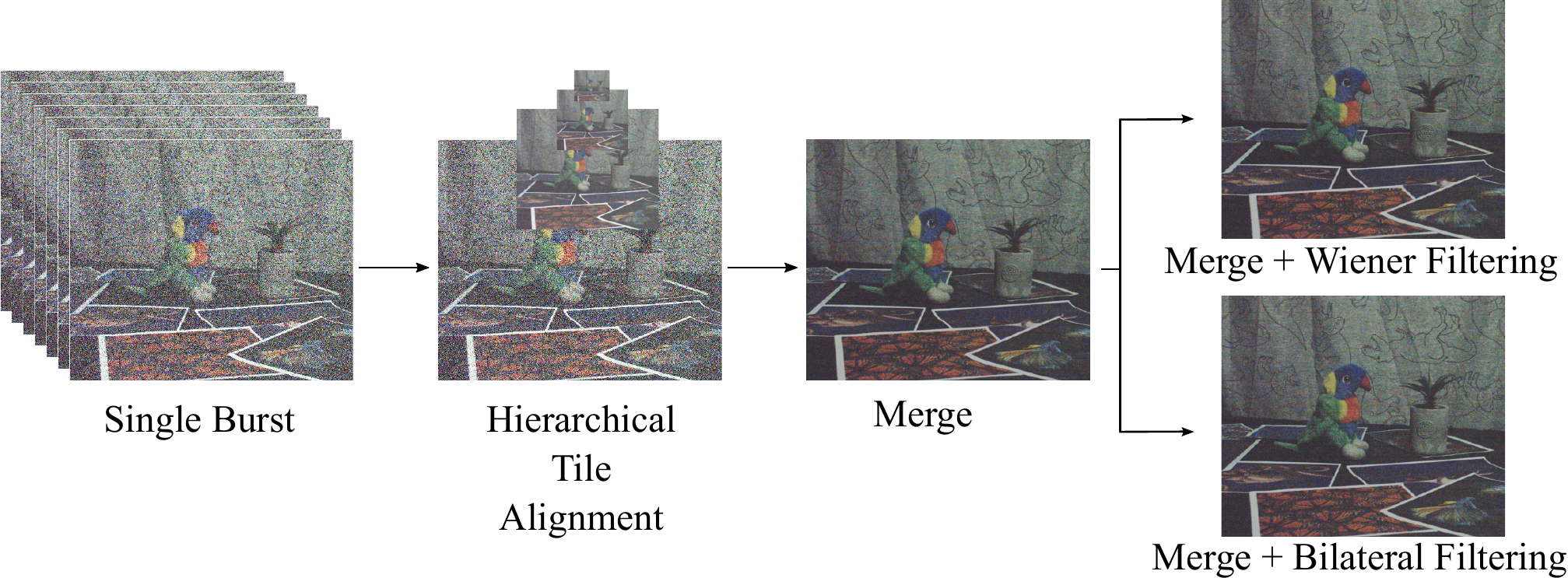}
	\caption{We improve the noise in the captured images by aligning every image in a burst with the chosen reference image of the burst. We merge the aligned images in temporal direction with a voting scheme to avoid misalignment. We use Wiener and bilateral filtering on temporally merged images. We give the outputs of this pipeline to COLMAP as three different inputs: burst with merge, burst with merge and Wiener filtering and burst with merge and bilateral filtering for reconstruction.}
	\label{fig_3}
\end{figure}
    
\subsection{Robust Temporal Merge}
We merge all the images in the burst together, following the pairwise method developed in~\cite{Hasinoff2016}. For a static scene, capturing and merging $N$ images ideally reduces random noise by $\sqrt{N}$. Merging method must also be robust to alignment failures that may occur due to changes in lighting, unaccounted scene motion at motion boundaries and occlusions.
    
To increase robustness to motion, we take the approach proposed in~\cite{Hasinoff2016} in which temporal filtering incorporates a controllable degree of contribution from each image:
    \begin{equation}
    \label{eq:merge}
    \overline{T_R}(\omega) = \frac{1}{N} \displaystyle \sum_{z=1}^{N}T_z(\omega) + A_z(\omega)[T_R(\omega) - T_z(\omega)].
    \end{equation}
$\overline{T_R}(\omega)$ is the temporally filtered tile in frequency domain, $T_z(\omega)$ is the computed 2D \gls{DFT} of the corresponding alternative tile and $T_R(\omega)$ is the computed 2D \gls{DFT} of the corresponding common tile. $z$ is the frame index and $N$ is the number of frames within a burst.
    
$A_z$ is the contribution of each frame, found as
    \begin{equation}
    \label{eq:vote}
    A_z(\omega) = \frac{|D_z(\omega)|^2}{|D_z(\omega)|^2 + c\sigma^2},
    \end{equation}
where $D_z(\omega)$ = $T_R(\omega)$ - $T_z(\omega)$. $\sigma^2$ is the noise variance and $c$ is the degree of contribution that increases noise reduction at the expense of robustness.
    
We have control over the degree of contribution $c$ and the tile size used in filtering for fine tuning merged images. Ringing artifacts that are commonly associated with frequency domain filters are avoided with a windowing approach while other distortions that are not caused by the noise in the images are treated as misalignment in the frequency domain by the robust filtering method. 
    
We also perform hybrid denoising by adding spatial filters separately on temporally merged images as depicted in Fig.~\ref{fig_3}. We filter high spatial frequency content aggressively to maximize subject quality with a noise shaping Wiener filter in 2D \gls{DFT} domain similar to~\cite{Hasinoff2016}. Separately, we apply bilateral filter as an edge-preserving filter on the temporally merged images in spatial domain.

\begin{figure}
	\centering
	\includegraphics[width=\columnwidth]{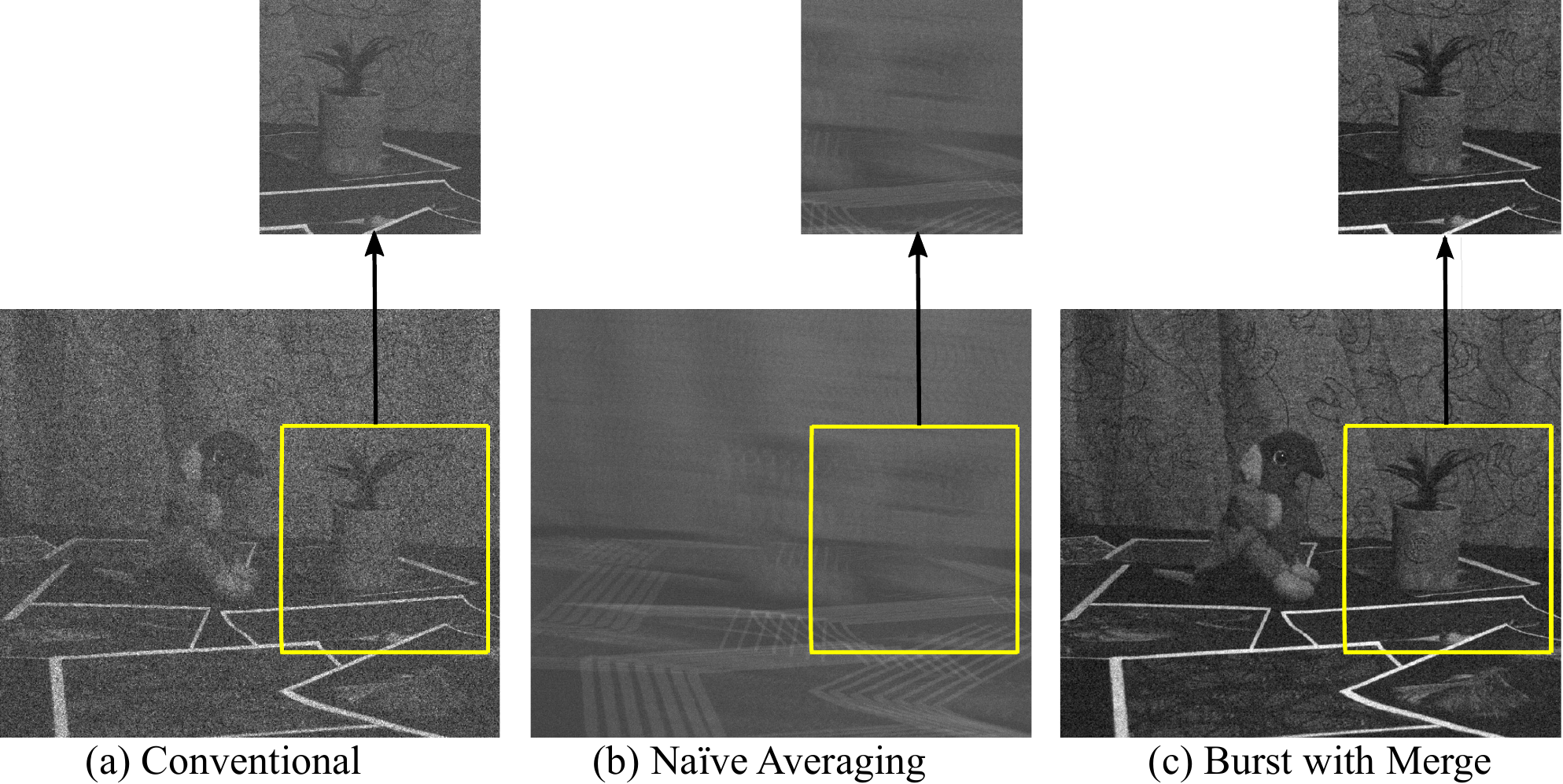}
	\caption{Application of alternative exposure scheme. (a) Single frame captured over an exposure time of $t$ ms. (b) Multiple frames captured over $T_a$ ms and merged na\"ively. This represents long exposure scheme with motion blur. (c) Our proposed merging approach}
	\label{fig_4}
\end{figure}

Fig.~\ref{fig_4} illustrates the performance of the align-and-merge approach, comparing with “conventional” noisy image, which corresponds to a single raw frame taken from a burst. Na\"ive averaging the images in a burst to a common image without alignment fails due to motion within the burst. This also demonstrates similarity towards capturing an image with longer exposure. Burst with merge demonstrates denoised image with higher \gls{SNR} after successful tile-based hierarchical alignment and robust temporal merge.

\subsection{Reconstruction Pipeline}
We use COLMAP~\cite{Schonberger2016}, an end-to-end feature-based state-of-the-art reconstruction pipeline to extract, match and geometrically verify features between sparse camera poses. We demonstrate sparse reconstruction by triangulating scene points and refining via bundle adjustment. We evaluate features following the procedures used in~\cite{Schonberger2017, Dansereau2019}.
\section{RESULTS}
\label{sec:results}

In the following we first evaluate improvements in feature detection and accuracy afforded by our method in noisy imagery using synthetic image sequences. Then we quantitatively evaluate our method in an \gls{SfM} pipeline, comparing against conventional image capture and direct use of all images in the burst. We consider both 3D reconstruction performance and camera trajectory accuracy.
    
\subsection{Feature Performance in Noise}
\begin{figure}
	\centering
	\includegraphics[width=\columnwidth]{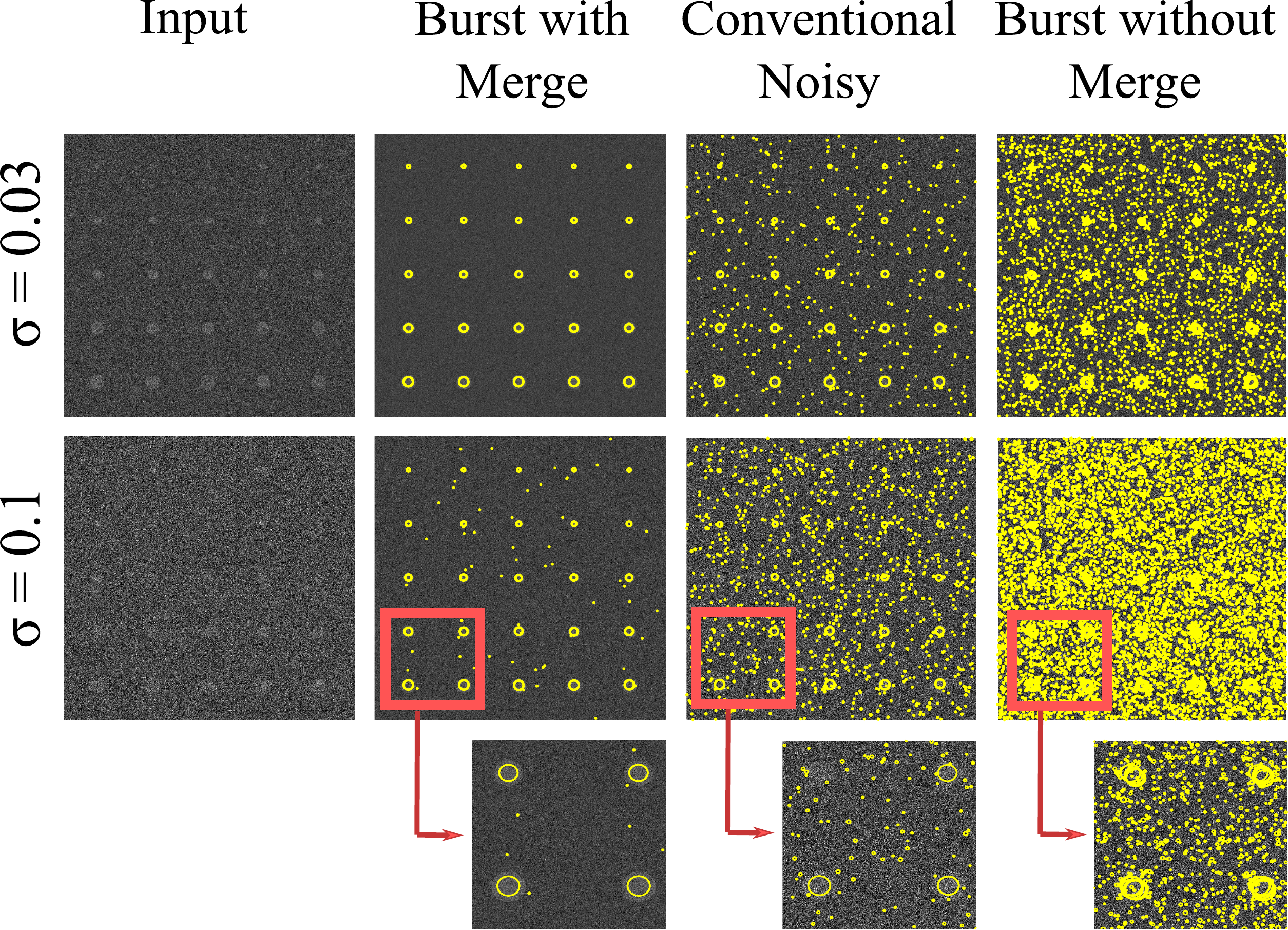}
	\caption{Detected features on synthetic images using \gls{SIFT}, presented at two noise levels $\sigma$. At the lower noise level (top row), our method performs well with no spurious features, while conventional noisy method and burst without merge generates more spurious features. At higher noise (bottom row), while our method detects all true features, conventional noisy approach detects more spurious features and less true features than ours and burst without merge registers overwhelming spurious features.}
	\label{fig_5}
\end{figure}

We use a synthetic set of scenes with known feature locations to demonstrate feature performance in different noise levels. The synthetic images, shown in Fig.~\ref{fig_5}, have 25 disks at varying scales. Every image has an apparent motion of 12 or less pixels between each other. The contrast between the disks and background is 0.1. We introduce moderate noise with variance 0.03 (top) and strong noise with variance 0.1 (bottom).
    
We extract \gls{SIFT} features at a peak threshold of 0.015. It is evident from Fig.~\ref{fig_5} that burst with merge performs better than burst without merge and conventional noisy with no spurious features for the moderate noise case ($\sigma$ = 0.03). While our method extracts all true features in strong noise, the conventional method fails to extract all true features and burst without merge is overwhelmed by spurious features.

\begin{figure}[t]
	\centering
	\includegraphics[width=\columnwidth]{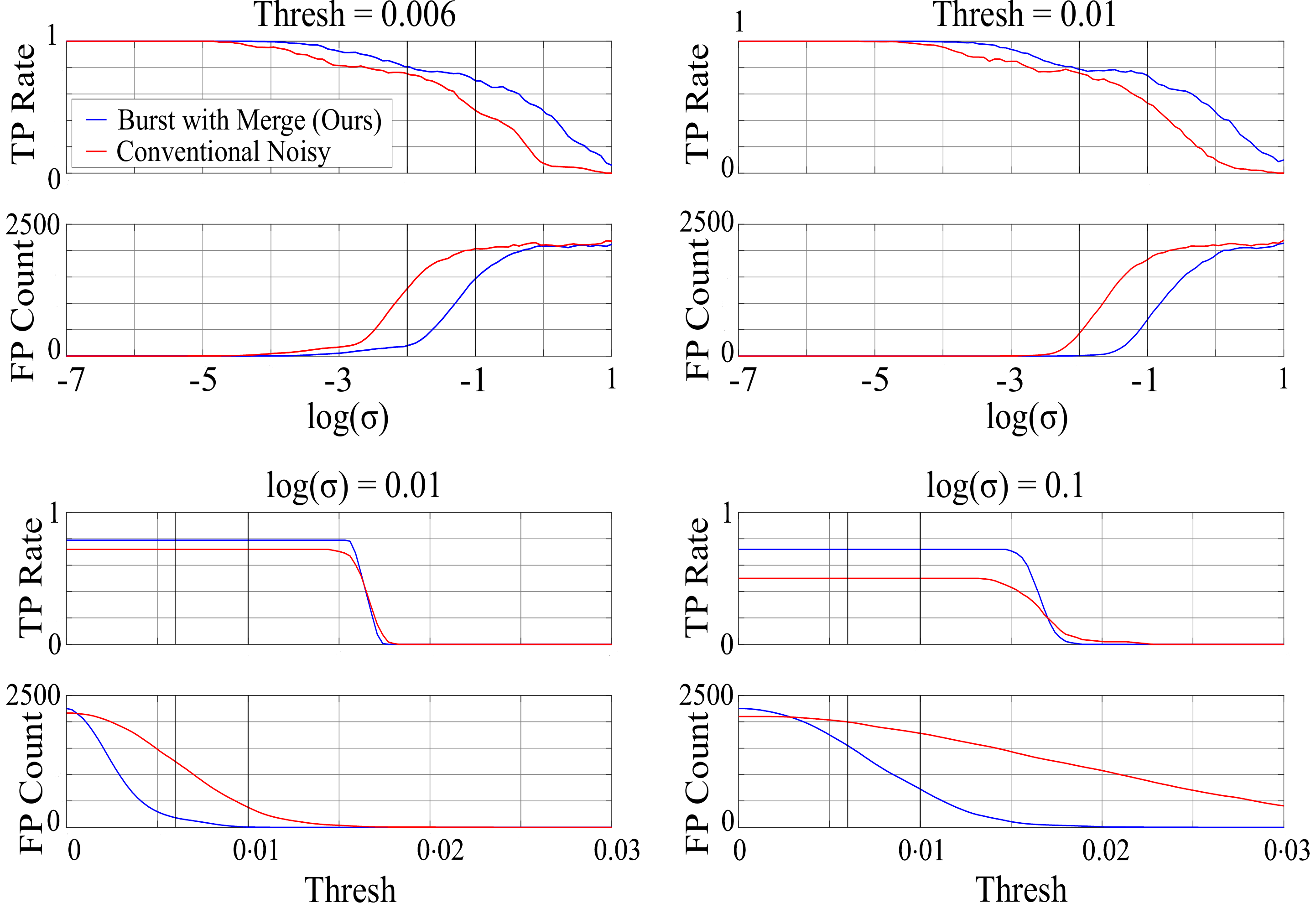}
	\caption{Noise performance: (top row) Sweeping noise level $\sigma$ for fixed detection thresholds, our method (red) shows a higher true positive (TP) rate and lower false positive (FP) count than conventional imaging (blue). (bottom row) Sweeping detection threshold, our method delivers a much higher TP rate in high noise and lower FP count for appropriately set threshold than conventional imaging. Overall, our method matches or outperforms conventional method in feature performance in noise.}
	\label{fig_6}
\end{figure}

We demonstrate the measurement of \gls{TP} rate and \gls{FP} count quantitatively on these synthetic images with varying noise levels and peak detection thresholds as shown in Fig.~\ref{fig_6} for both the proposed and conventional methods. The top row shows the \gls{TP} rate and \gls{FP} count at peak detection thresholds of 0.006 and 0.01 for a range of noise levels. 

An appropriate peak threshold gives negligible \gls{FP} count and higher \gls{TP} rate at lower noise level. As the amount of noise increases, our method outperforms the conventional method yielding more true positive features and less false positive count.
    
The bottom row of Fig.~\ref{fig_6} depicts the performance of both methods in two different noise levels for peak detection thresholds between 0.006 and 0.03.  Our method results in less false positive count and more true features. The lower false positive count results in lower overall computational requirements, as fewer putative matches need to be evaluated and rejected.

\begin{figure}
	\centering
	\includegraphics[width=60mm]{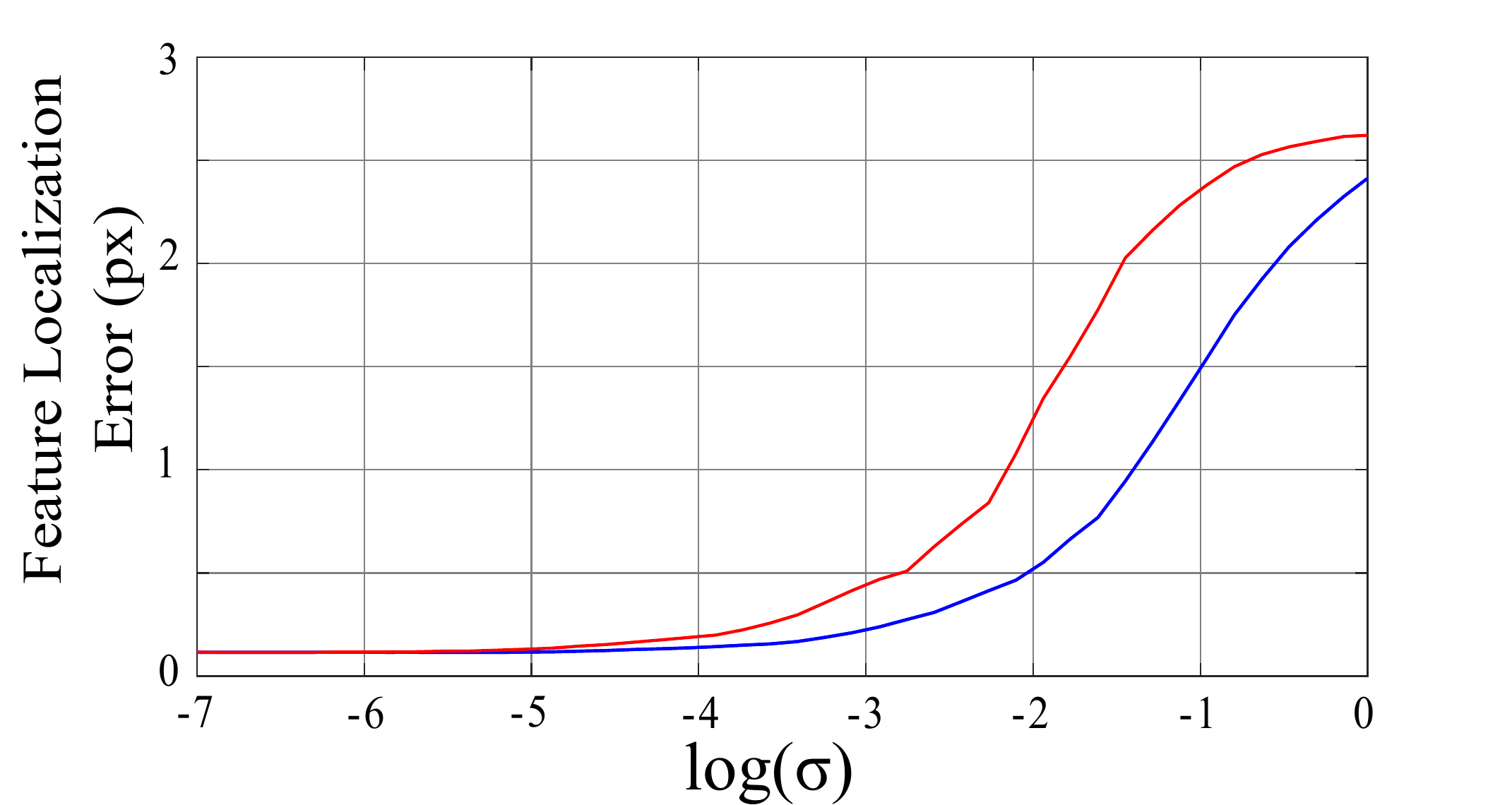}
	\caption{Feature localization accuracy: Sweeping noise level $\sigma$ for fixed detection threshold of 0.015, our method (blue) outperforms conventional imaging (red) by yielding accurate feature localization.}
	\label{fig_7}
\end{figure}

For a selected peak detection threshold of 0.015, we also evaluated the localization accuracy of the extracted true positive features compared to the conventional approach as depicted in Fig.~\ref{fig_7}. As noise level increases, our method outperforms the conventional approach by yielding significantly lower feature localisation errors.

\subsection{Reconstruction Accuracy}
We demonstrate our method by mounting a monocular machine vision camera FLIR FL3-U3-120S3C-C with an f/2.1 lens on a UR5e robotic arm as shown in Fig.~\ref{fig_1}. We capture 16 bit raw Bayer images of size 2992 x 2500. We capture 22 bursts of 7 images each for a single trajectory, and repeat the trajectory for 20 scenes composed of objects with different textures, shapes and sizes in an environment with controlled lighting. We capture our dataset at 1 ms and 0.1 ms exposure times, adjusting the gain in each case to maximize contrast while avoiding saturation.

We run COLMAP to generate sparse reconstructions of 20 different light-constrained scenes using our methods: burst with merge, burst with merge and Wiener filtering and burst with merge and bilateral filtering. We compare against alternative approaches: burst without merge where we directly use all images in the burst and conventional single-image capture, with the single image corresponding to the centre most frame of the burst, as depicted in Fig.~\ref{fig_2}.
    
We employ COLMAP with default settings, meaning a constant feature peak threshold of 0.0066 is employed for all images. To more fairly compare methods, we also repeat the experiment with peak thresholds adjusted to suit the different noise levels yielded by each method. Because unmerged images have more noise, they call for a more selective peak threshold. We empirically select peak thresholds of 0.001 and 0.005 for the proposed and conventional methods, respectively, as these yield similar levels of spurious feature detection.
    
Finally, we repeated the same experiments with COLMAP configured to be more permissive of images with few inlier feature matches, allowing as few as 15. This is useful when dealing with challenging scenes and allowed more images to be successfully incorporated in to the \gls{SfM} solution.

Following the feature comparison approach in~\cite{Schonberger2017}, we evaluate reconstruction performance as shown in \Table{\ref{tab:1}} in terms of numbers of keypoints per image, putative feature matches per image, number of putative matches classified as inliers, match ratio: the proportion of detected features yielding putative matches, precision: the proportion of putative matches yielding inlier matches, matching score: the proportion of detected features yielding inlier matches, the mean number of 3D points in the reconstructed models and the mean number of 3D points per image.
    
At a moderate noise level, i.e. for images captured over 1ms, our method outperforms alternative approaches across all metrics, by reconstructing all images passed for all scenes at default settings.
    
At a higher noise level, i.e: for images captured over 0.1ms, not all images are reconstructed with default settings, and thus, we evaluate by tuning peak threshold values. The proposed method successfully reconstructed all the scenes with the strongest putative matches per image, inlier matches per image, match ratio, match score and 3D points per image. Our method outperforms the conventional approach or shows competitive results across all metrics.
    
While burst without merge uses 154 images and takes 49.58 minutes on average for reconstruction of a scene, our method uses only 22 images and takes 3.25 minutes on average for reconstruction on an Intel i7-9700 at 4.70 GHz. We extract features using NVIDIA GTX770 for reconstruction. Our method also reduces the failure rate of reconstruction due to non-convergence and thus reconstructs more scenes than alternative approaches in strong noise.
    
\begin{table*}[t]
\centering
\vspace{0.3cm}
\caption{Average performance of reconstruction for light-constrained scenes. \gls{SfM}: with COLMAP’s default values, our approach outperforms captured images in all measures, including successful reconstruction of significantly more registered images per input scenes, more 3D points per image, more putative matches and inlier matches per match per image. By tuning COLMAP parameters, our approach perform successful reconstruction at extreme low light. The results are evaluated over 20 scenes. Bold: Best results, Red: Best/Second best results from conventional approaches; Green: Best/Second best results from our proposed approaches}
\label{tab:1}
\resizebox{\linewidth}{!}{%
\begin{tabular}{lllllllllllll} \hline
\begin{tabular}[c]{@{}l@{}}Reconstruction\\Modes\end{tabular} & Method & \begin{tabular}[c]{@{}l@{}}No. \\of\\images\end{tabular} & \begin{tabular}[c]{@{}l@{}}No. of\\images\\pass\end{tabular} & \begin{tabular}[c]{@{}l@{}}\% \\images\\pass\end{tabular} & \begin{tabular}[c]{@{}l@{}}Key\\Points/\\Image\end{tabular} & \begin{tabular}[c]{@{}l@{}}Putative\\Matches/\\Image\end{tabular} & \begin{tabular}[c]{@{}l@{}}Inlier \\Matches/\\Image\end{tabular} & \begin{tabular}[c]{@{}l@{}}Match\\Ratio\end{tabular} & Precision & \begin{tabular}[c]{@{}l@{}}Match\\Score\end{tabular} & \begin{tabular}[c]{@{}l@{}}3D\\Points\end{tabular} & \begin{tabular}[c]{@{}l@{}}3D \\Points/\\Image\end{tabular} \\ \hline
\multirow{5}{*}{\begin{tabular}[c]{@{}l@{}}1ms \\Default\end{tabular}} & Burst without merge & 154 & 152 & \textcolor{red}{98.70} & 9116 & 681 & 629 & 7.50E-02 & \textcolor{red}{0.92} & 6.90E-02 & 39790 & 258 \\
 & Conventional noisy & 22 & 21.6 & 98.18 & 9106 & 656 & 603 & 7.20E-02 & 0.92 & 6.60E-02 & 8270 & 376 \\ \cline{2-13}
 & Burst with merge & 22 & 22 & \textbf{\textcolor[rgb]{0,0.502,0}{100}} & 10522 & 866 & 801 & 8.20E-02 & \textbf{\textcolor[rgb]{0,0.502,0}{0.93}} & 7.60E-02 & 11130 & 506 \\
 & Burst with merge + Wiener & 22 & 22 & \textcolor[rgb]{0,0.502,0}{\textbf{100}} & 10300 & \textcolor[rgb]{0,0.502,0}{907} & \textcolor[rgb]{0,0.502,0}{842} & \textcolor[rgb]{0,0.502,0}{8.80E-02} & \textbf{\textcolor[rgb]{0,0.502,0}{0.93}} & \textcolor[rgb]{0,0.502,0}{8.20E-02} & 11387 & \textcolor[rgb]{0,0.502,0}{518} \\
 & Burst with merge + bilateral & 22 & 22 & \textcolor[rgb]{0,0.502,0}{\textbf{100}} & 10202 & \textbf{\textcolor[rgb]{0,0.502,0}{936}} & \textbf{\textcolor[rgb]{0,0.502,0}{868}} & \textbf{\textcolor[rgb]{0,0.502,0}{9.20E-02}} & \textbf{\textcolor[rgb]{0,0.502,0}{0.93}} & \textbf{\textcolor[rgb]{0,0.502,0}{8.50E-02}} & 11693 & \textbf{\textcolor[rgb]{0,0.502,0}{532}} \\ \hline
\multirow{5}{*}{\begin{tabular}[c]{@{}l@{}}0.1ms\\Default\end{tabular}} & Burst without merge & 154 & 89.85 & 58.34 & 6187 & 37 & 29 & \textcolor{red}{5.90E-03} & \textbf{\textcolor{red}{0.78}} & \textcolor{red}{4.60E-03} & 3456 & \textcolor{red}{22} \\
 & Conventional noisy & 22 & 13.5 & 61.36 & 6187 & 34 & 25 & 5.50E-03 & \textcolor{red}{0.74} & 4.00E-03 & 382 & 17 \\ \cline{2-13}
 & Burst with merge & 22 & 16.55 & \textbf{\textcolor[rgb]{0,0.502,0}{75}} & 9344 & 45 & 33 & 4.80E-03 & 0.73 & 3.50E-03 & 574 & \textbf{\textcolor[rgb]{0,0.502,0}{26}} \\
 & Burst with merge + Wiener & 22 & 15 & 68 & 10978 & \textcolor[rgb]{0,0.502,0}{47} & \textcolor[rgb]{0,0.502,0}{34} & 4.20E-03 & 0.72 & 3.00E-03 & 571 & \textbf{\textcolor[rgb]{0,0.502,0}{26}} \\
 & Burst with merge + bilateral & 22 & 15.8 & \textcolor[rgb]{0,0.502,0}{71.8} & 7368 & \textbf{\textcolor[rgb]{0,0.502,0}{50}} & \textbf{\textcolor[rgb]{0,0.502,0}{36}} & \textbf{\textcolor[rgb]{0,0.502,0}{6.70E-03}} & 0.72 & \textbf{\textcolor[rgb]{0,0.502,0}{4.90E-03}} & 566 & \textbf{\textcolor[rgb]{0,0.502,0}{26}} \\ \hline
\multirow{5}{*}{\begin{tabular}[c]{@{}l@{}}0.1ms \\Permissive \\Matching\end{tabular}} & Burst without merge & 154 & 76.9 & 49.94 & 6187 & 37 & 29 & \textcolor{red}{5.90E-03} & \textcolor{red}{\textbf{0.78}} & \textcolor{red}{4.60E-03} & 2623 & 17 \\
 & Conventional noisy & 22 & 16.85 & 76.59 & 6187 & 34 & 25 & 5.50E-03 & \textcolor{red}{0.74} & 4.00E-03 & 422 & 19 \\ \cline{2-13}
 & Burst with merge & 22 & 21 & 95.45 & 9344 & 45 & 33 & 4.80E-03 & 0.73 & 3.50E-03 & 718 & 34 \\
 & Burst with merge + Wiener & 22 & 21.05 & \textcolor[rgb]{0,0.502,0}{95.68} & 10978 & \textcolor[rgb]{0,0.502,0}{47} & \textcolor[rgb]{0,0.502,0}{34} & 4.20E-03 & 0.72 & 3.00E-03 & 781 & \textbf{\textcolor[rgb]{0,0.502,0}{36}} \\
 & Burst with merge + bilateral & 22 & 21.9 & \textbf{\textcolor[rgb]{0,0.502,0}{99.55}} & 7368 & \textbf{\textcolor[rgb]{0,0.502,0}{50}} & \textbf{\textcolor[rgb]{0,0.502,0}{36}} & \textbf{\textcolor[rgb]{0,0.502,0}{6.70E-03}} & 0.72 & \textbf{\textcolor[rgb]{0,0.502,0}{4.90E-03}} & 780 & \textcolor[rgb]{0,0.502,0}{35} \\ \hline
\multirow{5}{*}{\begin{tabular}[c]{@{}l@{}}0.1ms Peak \\Threshold\\Tuned\end{tabular}} & Burst without merge & 154 & 85.95 & 55.81 & 5431 & 43 & 33 & 7.90E-03 & \textcolor{red}{\textbf{0.77}} & 6.10E-03 & 3826 & 25 \\
 & Conventional noisy & 22 & 16.7 & 75.9 & 5431 & 39 & 30 & 7.00E-03 & \textbf{\textcolor{red}{0.77}} & 5.50E-03 & 536 & 24 \\ \cline{2-13}
 & Burst with merge & 22 & 20.9 & \textbf{\textcolor[rgb]{0,0.502,0}{95}} & 5367 & 50 & 36 & \textbf{\textcolor[rgb]{0,0.502,0}{9.30E-03}} & 0.72 & \textbf{\textcolor[rgb]{0,0.502,0}{6.70E-03}} & 852 & \textbf{\textcolor[rgb]{0,0.502,0}{39}} \\
 & Burst with merge + Wiener & 22 & 18.45 & 83 & 5977 & \textcolor[rgb]{0,0.502,0}{51} & \textcolor[rgb]{0,0.502,0}{37} & \textcolor[rgb]{0,0.502,0}{8.50E-03} & \textcolor[rgb]{0,0.502,0}{0.73} & \textcolor[rgb]{0,0.502,0}{6.20E-03} & 792 & \textcolor[rgb]{0,0.502,0}{36} \\
 & Burst with merge + bilateral & 22 & 18.7 & \textcolor[rgb]{0,0.502,0}{85} & 7572 & \textbf{\textcolor[rgb]{0,0.502,0}{53}} & \textbf{\textcolor[rgb]{0,0.502,0}{38}} & 6.90E-03 & 0.72 & 5.00E-03 & 731 & 33 \\ \hline
\multirow{5}{*}{\begin{tabular}[c]{@{}l@{}}0.1ms Peak\\Threshold\\Tuned +\\Permissive\\Mapping\end{tabular}} & Burst without merge & 154 & 50.7 & 32.92 & 5431 & 43 & 33 & 7.90E-03 & \textcolor{red}{\textbf{0.77}} & 6.10E-03 & 2152 & 14 \\
 & Conventional noisy & 22 & 19.3 & 87.73 & 5431 & 39 & 30 & 7.00E-03 & \textbf{\textcolor{red}{0.77}} & 5.50E-03 & 576 & 26 \\ \cline{2-13}
 & Burst with merge & 22 & 22 & \textbf{\textcolor[rgb]{0,0.502,0}{100}} & 5367 & 50 & 36 & \textbf{\textcolor[rgb]{0,0.502,0}{9.30E-03}} & 0.72 & \textbf{\textcolor[rgb]{0,0.502,0}{6.70E-03}} & 864 & \textcolor[rgb]{0,0.502,0}{39} \\
 & Burst with merge + Wiener & 22 & 21.9 & \textcolor[rgb]{0,0.502,0}{99.55} & 5977 & \textcolor[rgb]{0,0.502,0}{51} & \textcolor[rgb]{0,0.502,0}{37} & \textcolor[rgb]{0,0.502,0}{8.50E-03} & \textcolor[rgb]{0,0.502,0}{0.73} & \textcolor[rgb]{0,0.502,0}{6.20E-03} & 848 & \textcolor[rgb]{0,0.502,0}{39} \\
 & Burst with merge + bilateral & 22 & 22 & \textbf{\textcolor[rgb]{0,0.502,0}{100}} & 7572 & \textbf{\textcolor[rgb]{0,0.502,0}{53}} & \textbf{\textcolor[rgb]{0,0.502,0}{38}} & 6.90E-03 & 0.72 & 5.00E-03 & 886 & \textcolor[rgb]{0,0.502,0}{\textbf{40}} \\ \hline
\end{tabular}
}
\end{table*}

\subsection{Camera Trajectory Accuracy}
We evaluate the accuracy of the camera trajectory estimate by using a robotic arm to collect ground truth poses. We aligned the camera poses to the ground truth poses as there is an arbitrary scale factor involved in monocular \gls{SfM}. The arbitrary scale we use when reporting results is determined by the distance between the first pair of registered images. Fig.~\ref{fig_8} shows how our method reconstruct more accurate camera poses than the competing method. We compute absolute instantaneous error and relative pose error between reconstructed camera poses and ground truth poses for translation and rotation as shown in \Table{\ref{tab:2}}. Our method is competitive with moderate noise and outperforms alternative approaches with strong noise.

\begin{table*}[t]
\centering
\caption{Mean translation error and mean rotational error in camera poses for all reconstruction methods: Our approach outperforms conventional noisy approach with same number of images and performs competitive with burst without merge which has seven times more images than our approach. Bold: Best results, Red: Best/Second best results from conventional approaches; Green: Best/Second best results from our proposed approaches}
\label{tab:2}
\begin{tabular}{llllllllllll} 
\hline
\multirow{3}{*}{\begin{tabular}[c]{@{}l@{}}Trajectory\\Evaluation\end{tabular}} & \multirow{3}{*}{\begin{tabular}[c]{@{}l@{}}Reconstruction\\Modes\end{tabular}} & \multicolumn{2}{l}{\begin{tabular}[c]{@{}l@{}}Burst without\\Merge\end{tabular}} & \multicolumn{2}{l}{\begin{tabular}[c]{@{}l@{}}Conventional\\Noisy\end{tabular}} & \multicolumn{2}{l}{\begin{tabular}[c]{@{}l@{}}Burst with\\Merge\end{tabular}} & \multicolumn{2}{l}{\begin{tabular}[c]{@{}l@{}}Burst with\\Merge + Wiener\end{tabular}} & \multicolumn{2}{l}{\begin{tabular}[c]{@{}l@{}}Burst with\\Merge + Bilateral\end{tabular}} \\ 
\cline{3-12}
 &  & trans. & rot. & trans. & rot. & trans. & rot. & trans. & rot. & trans. & rot. \\
 &  & (cm) & (deg) & (cm) & (deg) & (cm) & (deg) & (cm) & (deg) & (cm) & (deg) \\ 
\hline
\multirow{5}{*}{\begin{tabular}[c]{@{}l@{}}Absolute\\Instantaneous\\Error\end{tabular}} & Default (1ms) & \textbf{\textcolor{red}{0.73}} & 0.459 & 0.82 & 0.505 & \textcolor[rgb]{0,0.502,0}{0.74} & 0.452 & 0.79 & \textcolor[rgb]{0,0.502,0}{\textbf{0.446}} & 0.79 & \textcolor[rgb]{0,0.502,0}{0.451} \\
 & Default (0.1ms) & 2.83 & 0.715 & 4.51 & 0.775 & \textcolor[rgb]{0,0.502,0}{\textbf{2.42}} & \textbf{\textcolor[rgb]{0,0.502,0}{0.630}} & 2.82 & \textcolor[rgb]{0,0.502,0}{0.646} & \textcolor[rgb]{0,0.502,0}{2.74} & 0.757 \\
 & Permissive Mapping & 2.77 & \textbf{\textcolor{red}{0.499}} & 5.24 & 1.040 & \textcolor[rgb]{0,0.502,0}{2.25} & 0.662 & \textcolor[rgb]{0,0.502,0}{\textbf{2.10}} & \textcolor[rgb]{0,0.502,0}{0.512} & 2.81 & 0.645 \\
 & Peak Threshold & 2.46 & 0.771 & 5.43 & 1.155 & \textcolor[rgb]{0,0.502,0}{0.98} & \textbf{\textcolor[rgb]{0,0.502,0}{0.341}} & \textcolor[rgb]{0,0.502,0}{\textbf{0.89}} & \textcolor[rgb]{0,0.502,0}{0.421} & 2.04 & 0.530 \\
 & \begin{tabular}[c]{@{}l@{}}Peak Threshold +\\Permissive Mapping\end{tabular} & 2.75 & 0.487 & 2.92 & 0.615 & \textcolor[rgb]{0,0.502,0}{2.38} & \textcolor[rgb]{0,0.502,0}{0.451} & 2.70 & 0.473 & \textcolor[rgb]{0,0.502,0}{\textbf{1.49}} & \textcolor[rgb]{0,0.502,0}{\textbf{0.440}} \\ 
\hline
\multirow{5}{*}{\begin{tabular}[c]{@{}l@{}}Relative\\Pose\\Error\end{tabular}} & Default (1ms) & 2.70 & \textcolor{red}{\textbf{0.004}} & 2.25 & 0.005 & \textcolor[rgb]{0,0.502,0}{\textbf{1.99}} & \textcolor[rgb]{0,0.502,0}{\textbf{0.004}} & \textcolor[rgb]{0,0.502,0}{\textbf{2.00}} & \textcolor[rgb]{0,0.502,0}{\textbf{0.004}} & 2.64 & \textcolor[rgb]{0,0.502,0}{\textbf{0.004}} \\
 & Default (0.1ms) & 4.28 & 0.052 & \textcolor{red}{3.9} & 0.032 & \textcolor[rgb]{0,0.502,0}{\textbf{3.80}} & 0.030 & 4.27 & \textcolor[rgb]{0,0.502,0}{\textbf{0.015}} & \textcolor[rgb]{0,0.502,0}{3.87} & \textcolor[rgb]{0,0.502,0}{0.028} \\
 & Permissive Mapping & 4.99 & 0.035 & 7.15 & 0.126 & \textcolor[rgb]{0,0.502,0}{\textbf{3.16}} & \textcolor[rgb]{0,0.502,0}{\textbf{0.016}} & \textcolor[rgb]{0,0.502,0}{3.22} & \textbf{\textcolor[rgb]{0,0.502,0}{0.016}} & 4.24 & 0.028 \\
 & Peak Threshold & 3.36 & 0.021 & 5.66 & 0.044 & \textcolor[rgb]{0,0.502,0}{\textbf{1.99}} & \textbf{\textcolor[rgb]{0,0.502,0}{0.013}} & \textcolor[rgb]{0,0.502,0}{2.29} & \textcolor[rgb]{0,0.502,0}{\textbf{0.013}} & 2.53 & 0.021 \\
 & \begin{tabular}[c]{@{}l@{}}Peak Threshold +\\Permissive Mapping\end{tabular} & 4.66 & 0.039 & 4.71 & 0.091 & \textcolor[rgb]{0,0.502,0}{\textbf{2.62}} & \textcolor[rgb]{0,0.502,0}{0.018} & 3.96 & 0.021 & \textcolor[rgb]{0,0.502,0}{2.71} & \textcolor[rgb]{0,0.502,0}{0.017} \\
\hline
\end{tabular}
\end{table*}
\section{DISCUSSION}
\label{sec:disc}
Our key finding is that using sparse higher \gls{SNR} merged images is better than using lower \gls{SNR} unmerged images for light-constrained \gls{SfM}. We recommend using burst with merge for light-constrained SfM to increase feature performance, camera trajectory accuracy, and rate of success in incorporating challenging low-light images into the \gls{SfM} solution. Although our approach involves increased computation in merging images within a burst, this is offset by the decreased rate of false positive feature detection, reducing the burden of extraneous feature matching and rejection.

In adapting burst photography for reconstruction in low light, there exists many trade-offs between the number of images within a single burst, gain, exposure time, motion blur, speed of the robot, image pyramid designs for motion alignment and noise in the image. In the following, we provide a set of recommendation on how to use and tune multiple bursts with merging in reconstruction tasks for common robotics applications:

 \begin{figure}[t]
	\centering
	\includegraphics[width=\columnwidth]{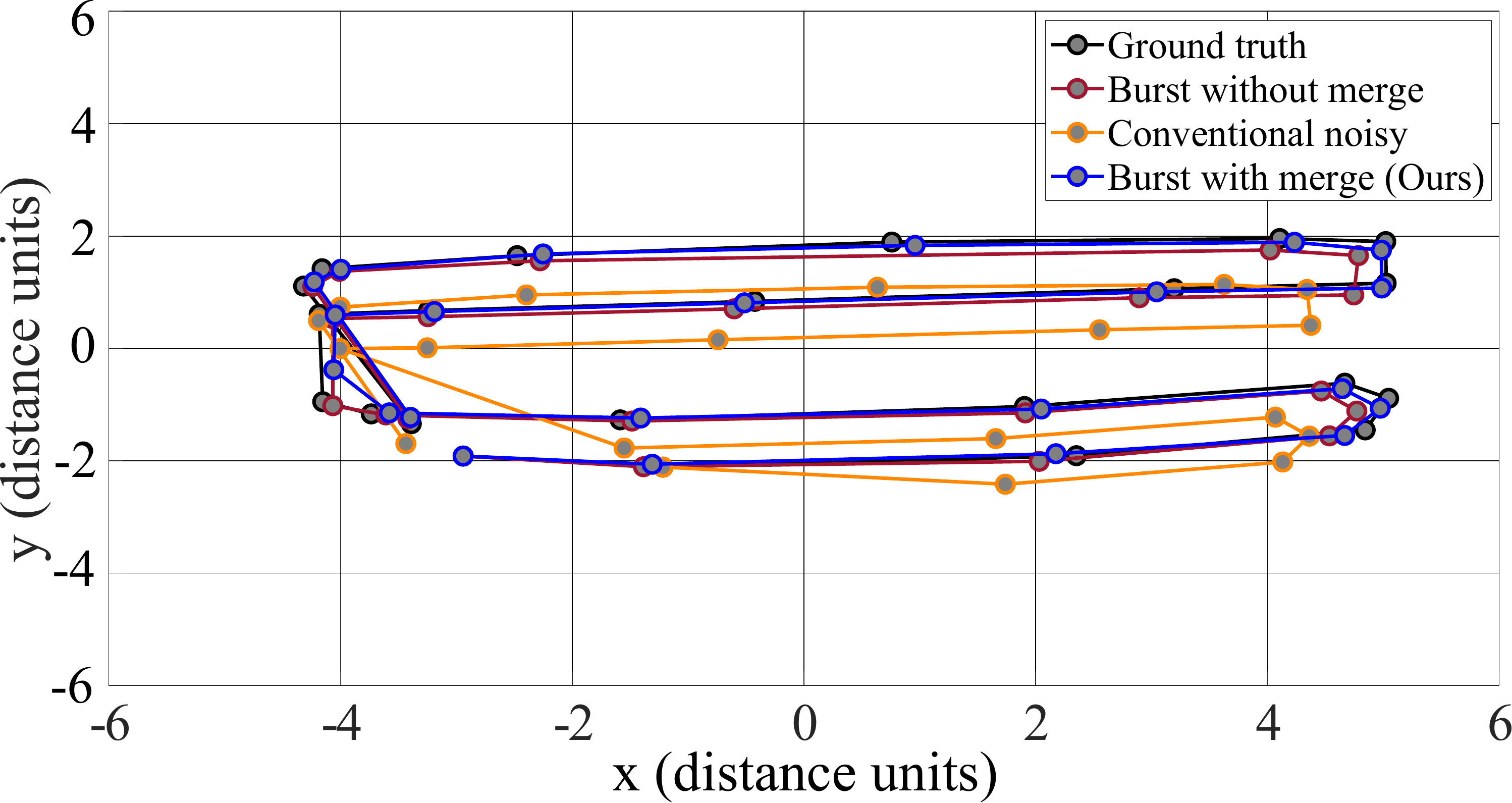}
    \caption{Camera trajectory for a particular captured scene (scene 16) in distance units, a measure between the first pair of registered images in a reconstruction. Our results reconstruct accurate camera trajectories using all input images; alternative approach that uses reconstructed 94.2\% of input noisy images without merge and conventional noisy approach that uses 95.5\% produces less accurate trajectory estimates.}
	\label{fig_8}
\end{figure}

\begin{itemize}
    \item Increase exposure time as high as possible without getting deleterious motion blur. This yields higher-\gls{SNR} input images and ultimately better reconstruction performance.
    \item Fixed-pattern noise can easily become the dominant source of noise in capturing low light images. This can be addressed by subtracting an average of multiple dark frames taken with the same gain, sensor temperature, and exposure time as the intended images within a burst. 
    \item Increase gain as high as possible without yielding excessive saturation. This amplifies both signal and noise, but is important in overcoming the  quantisation limit of the camera.
    \item Maximize the number of images in each burst to maximise \gls{SNR} in the merged images. The image count is ultimately limited by availability of compute, and the rate of apparent motion of the scene. Motion relative to the reference frame should not exceed 1/2 of the total frame. For this reason we also recommend minimizing the maximum apparent motion relative to the reference frame by employing the centre most pose as the reference.
    \item Maximize the number of pyramid level and minimize the search window size to avoid search getting trapped in local minimum for images with strong noise. We recommend having the minimum pyramid level higher than 3 and image size higher than 1/16 of the original image at the coarse level of the pyramid for robust alignment.
    \item For images with motion relative to the reference frame larger than 1/4 of the total frame, minimize the image size at each pyramid level not lower than 1/16 of the original image and maximize search extend to locate corresponding pixel coordinates of aligned tiles not higher than 32 pixels for robust alignment.
\end{itemize}
\section{CONCLUSIONS}
\label{sec:concl}

We adapted burst photography, commonly used in mobile photography for reconstruction in low light. We enabled the use of direct methods for image registration within the burst and used feature-based \gls{SfM} to handle the sparsity between bursts for reconstruction. We demonstrated successful reconstruction with decreased failure cases due to non-convergence compared to conventional methods. We also demonstrated improved performance relative to conventional imaging in true features, spurious features, putative matches, inlier matches, 3D points per images and accurate camera pose estimation.
    
Our method showed faster reconstruction with lower overall computational requirements compared to conventional methods. We expect that in more challenging low light conditions our method can improve the performance of 3D reconstruction and expand the range in which feature-based reconstruction can be applied.
    
This work is a first step toward solving the problem of 3D reconstruction in low light. For future work we anticipate employing adaptive sampling schemes, in which the parameters of burst capture and processing are dynamically chosen to suit the situation. We also expect the fusion of complementary sensors to yield interesting results.

%

{\small
	\bibliographystyle{IEEEtran}
	\bibliography{./references}
}
\end{document}